# Shapechanger: Environments for Transfer Learning

Sébastien Arnold    Tsam Kiu Pun    Théo-Tim J. Denisart    Francisco J. Valero-Cuevas

*Abstract*— We present *Shapechanger*, a library for transfer reinforcement learning specifically designed for robotic tasks. We consider three types of knowledge transfer—from simulation to simulation, from simulation to real, and from real to real—and a wide range of tasks with continuous states and actions. Shapechanger is under active development and open-sourced at: `https://github.com/seba-1511/shapechanger/`.

## I. Introduction: Challenges of Real-World RL

Reinforcement learning (RL) requires large amounts of data to properly train an agent. Often, when dealing with real-world systems, this large amount of data is costly to acquire or simply not accessible. RL involves the interaction of a learning agent with an environment, be it in simulation or in the real world. The environment is often considered to be completely specified by a Markov Decision Process (MDP) whose states can be fully or partially observable. The overarching goal of RL is for the agent to learn an optimal policy $\pi^\star$ which maximizes the expected discounted reward throughout episodes in the MDP.

Classical algorithmic approaches to discovering the policy $\pi^\star$ are based on learning value functions

- $V(s)$ which estimates the expected reward for state $s$, and
- $Q(s, a)$ which estimates the expected reward for taking aciton $a$ in state $s$

It is known that exact solutions to RL problems are achievable through tabular methods, and convergence guarantees exist under mild assumptions. Unfortunately tabular methods are not suited to deal with continuous states and actions, and scale poorly with respect to the size and complexity of the environment. When dealing with real-world problems it is customary to use approximations in the value functions, hoping that they will apply to the real environment. In addition, policy gradient methods can also directly approximate the policy and thus by-pass the need for value functions.

While numerous recent efforts were quite successful using non-linear approximators to work with continuous and high-dimensional action/state spaces [1], they all still require large amounts of training data. This is a major issue when dealing with real-world applications where accurate simulators are unavailable.

Transfer learning offers a solution to the training data problem. It allows the use of previously acquired knowledge to be transferred to a new system/environment to speed up RL.

In this framework, the agent first learns a policy on a similar, but more convenient environment (e.g., a simulation or a an easier task), and then transfers the acquired knowledge to solve the desired task. Several studies have successfully applied this approach to robotics [2], and we believe in its promise to deliver fast-learning, multi-task agents.

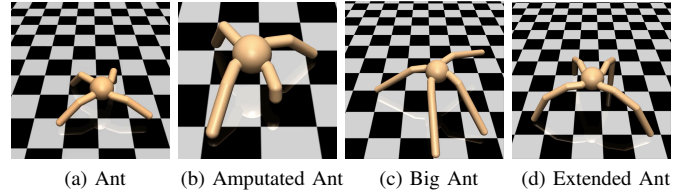

(a) Ant    (b) Amputated Ant    (c) Big Ant    (d) Extended Ant

Fig. 1: Modified Ants Environments

Regarding the availability of training environments, a major contribution was the open-source release of OpenAI Gym [3] and Universe. Thus, thousands of widely different environments are now available for the broader research community to benchmark and train their own agents. While environments are categorized by domain similarities (e.g. Atari, MuJoCo, Safety), none were specifically designed for transfer learning. Our work extends Gym in that direction by providing modified and new environments to its collection.

## II. The Shapechanger Environments

The goal of this paper is to offer three classes of environments to study and evaluate algorithms for transfer learning: Simulation to Simulation, Simulation to Real, and Real to Real.

### A. Class 1: Simulation to Simulation

Our Simulation to Simulation environments are implemented on top of the MuJoCo Physics engine [4]. The current version includes five extensions to existing environments in OpenAI Gym. Two of them are based on `InvertedPendulum-v1`, and simply rescale the length of the pendulum by factor of 2. (Shorter and longer). The other three are based on `Ant-v1` and respectively chop, lengthen, or add joints to the ant's limbs.

Most recent continuous control algorithms are able to solve the pendulum-based environments in a matter of minutes. Hence their principal use is for debugging and fast prototyping of new transfer techniques. On the other hand, the ants environments prove to be quite challenging. The Big Ant has a very similar setup to the original Ant, and an optimal policy on one should work fairly well on the other too. This is not true of the Amputated Ant; the optimal policy seems to completely ignore the damaged limb. Finally, we included a task with different action-state dimensionality, the Extended Ant. Depictions of all Ants environments are displayed in Figure 1.

*B. Class 2: Simulation to Real*

Simulation to Real transfer environments are arguably the most challenging to design. Since on-board robot learning remains a challenging task, precise and accurate knowledge of real physics are required to obtain meaningful knowledge transfer.

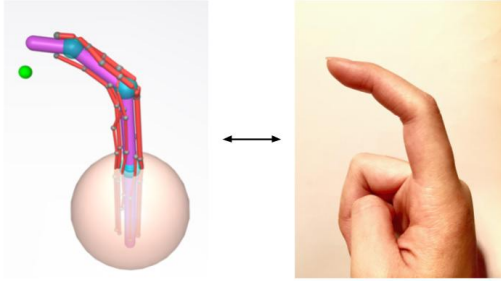

Fig. 2: Transfer from `Finger-v1` simulation to a real finger

Shapechanger currently only offers a single simulation-to-robot transfer environment: `Finger-v1`. Finger-v1 is an anatomically-inspired implementation of a human tendon-driven index finger. It is also built on top of MuJoCo, and similar in spirit to `Reacher-v1`, where the goal is to bring the tip of the finger close to a randomly sampled coordinate in 3D space. Once mastered, the agent should have gained an approximate understanding of the physics of human fingers which could then be transfered for the control of actual human fingers. [5 – 6] This transfer can be seen in Figure 2. The same finger model can also be used for learning dynamic movements. In this case, the aim of the agent is to control the finger such that its tip follows a moving target whose path can be specified a priori.

*C. Class 3: Real to Real*

Transferring knowledge from robot to robot is an important yet difficult task. Our preliminary environments are all vehicle-based and abstract the dynamics through steering and propulsive actuators. The state is a forward-facing camera image, and the goal of the agent is to avoid collisions while maximizing velocity. The transfer goal for all environments is to learn the physics of a vehicle and transfer that knowledge to the control of other vehicles.

Our first environment is a modified toy remote-controlled car, pictured in left of Figure 3. We replaced the factory micro-controller with a Raspberry Pi connected to a WiFi network, and programmed the GPIO ports to control the motors. The vehicle can be controlled through a Gym environment on a distinct computational node which communicates with the Raspberry Pi via TCP sockets. A live camera-feed and human manual controls are available, allowing the agent to learn by demonstration.

Because of our chosen abstraction, the same environment can be used to learn the control of different vehicles. As such the development of a boat, a tank, and a truck only required us to map the GPIO channels to the corresponding motor interfaces.

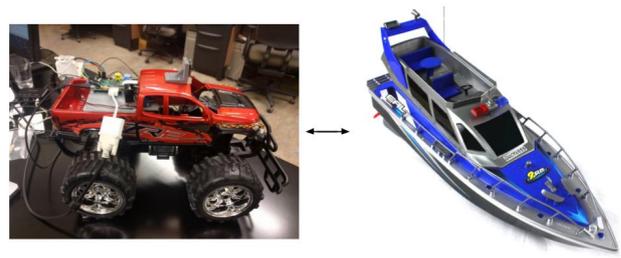

Fig. 3: Real Environment Transfer

## III. CONCLUSION

We presented Shapechanger, a library of environments for transfer learning in the context of RL. We compartmentalized the overarching task of transfer learning into three classes, which help overcome the difficulty in handling real-world tasks in various environments. Shapechanger is a continuing effort that regularly adds new environments to its environment library.

## IV. ACKNOWLEDGEMENTS

The authors would like to thank Elizabeth Chu for her help with MuJoCo, and her contribution to the tendon-driven finger.

This project is supported by NIH NIAMS R01 AR050520 and R01 AR052345 to FVC.